\documentclass{ifacconf}

\makeatletter
\let\old@ssect\@ssect 
\makeatother

\usepackage{graphicx}      
\usepackage{natbib}        
\usepackage{amsmath,amssymb,amsfonts}
\DeclareMathOperator*{\argmin}{arg\,min}
\usepackage{subcaption}
\usepackage{color}
\usepackage{hyperref}

\makeatletter
\def\@ssect#1#2#3#4#5#6{%
  \NR@gettitle{#6}
  \old@ssect{#1}{#2}{#3}{#4}{#5}{#6}
}
\makeatother

\begin{document}
\begin{frontmatter}

\title{Event Camera and LiDAR based Human Tracking for Adverse Lighting Conditions in Subterranean Environments \thanksref{footnoteinfo}}

\thanks[footnoteinfo]{We recommend the reader to watch the video of the proposed framework at \url{https://youtu.be/kIJ61VyIVTM}}
\thanks[footnoteinfo]{This work has been partially funded by the European Unions Horizon 2020 Research and Innovation Programme under the Grant Agreement No. 101003591 NEX-GEN SIMS.}


 \author[First]{Mario A.V. Saucedo},
 \author[First]{Akash Patel},
 \author[First]{Rucha Sawlekar},
 \author[First]{Akshit Saradagi},
 \author[First]{Christoforos Kanellakis}, 
 \author[Second]{Ali-Akbar Agha-Mohammadi} and
 \author[First]{George Nikolakopoulos}


\address[First]{Robotics and AI Team, Luleå University of Technology, Luleå, SE-971 87, Sweden, (e-mail: marval@ltu.se).}
\address[Second]{AI for Humanity Inc.}

\begin{abstract}  
In this article, we propose a novel LiDAR and event camera fusion modality for subterranean (SubT) environments for fast and precise object and human detection in a wide variety of adverse lighting conditions, such as low or no light, high-contrast zones and in the presence of blinding light sources. In the proposed approach, information from the event camera and LiDAR are fused to localize a human or an object-of-interest in a robot's local frame. The local detection is then transformed into the inertial frame and used to set references for a Nonlinear Model Predictive Controller (NMPC) for reactive tracking of humans or objects in SubT environments. The proposed novel fusion uses intensity filtering and K-means clustering on the LiDAR point cloud and frequency filtering and connectivity clustering on the events induced in an event camera by the returning LiDAR beams. The centroids of the clusters in the event camera and LiDAR streams are then paired to localize reflective markers present on safety vests and signs in SubT environments. The efficacy of the proposed scheme has been experimentally validated in a real SubT environment (a mine) with a Pioneer 3AT mobile robot. The experimental results show real-time performance for human detection and the NMPC-based controller allows for reactive tracking of a human or object of interest, even in complete darkness.
\end{abstract}
\begin{keyword}
Event-based vision, Event camera and LiDAR fusion, Human detection and tracking, NMPC-based tracking.
\end{keyword}
\end{frontmatter}
\section{Introduction}
Subterranean (SubT) environments have become a point of interest and challenge for the robotics community in the recent years, partly due to the DARPA subterranean Challenge held in 2021. During the challenge, many universities and organizations had the opportunity to deploy state-of-the-art (SOTA) systems and solutions for SubT navigation and exploration. SubT environments presented 
numerous unique challenges 
for autonomous robotics, from autonomous navigation and localization in GPS denied environments (\cite{kanellakis2018towards}) to perception in adverse lighting conditions (\cite{palieri2020locus}). The task of object and human detection was of central importance in the challenge, where the robotic platforms were required to identify and localize a series of \textit{artifacts} and \textit{human survivors}. In this article, we present an event camera and LiDAR fusion framework for object and human detection and demonstrate the ability of the proposed framework to counter adverse and varied lighting conditions presented by SubT environments. 
\subsection{Related work}
Due to the presence of low-illumination and GPS unavailability, sensors like the LiDAR have become a popular choice for perception in SubT environments (\cite{jfr}). Although the advantages of LiDARs are undeniable, there are limitations to its use for object detection due to: 1) the well-known limitations of the heavy computational cost of point cloud-based detection algorithms, 2) the presence of complex solid formations that leads to missed detections, 3) the point clouds become increasingly sparse with distance, which leads to loss of data and 4) difficulty in the detection of non-3D shapes like reflective signs or markers in SubT environments. For these reasons, in many cases, it is common to pair a LiDAR with light and inexpensive sensors, such as RGB Cameras. In SubT environments, where lighting conditions are inconsistent and defiant even for human eyes, the use of RGB cameras becomes difficult and presents several complexities. In many cases, it requires the aid of illumination hardware, which does not entirely resolve the low-light or high-contrast conditions observed in most SubT scenarios. Furthermore, if the system in question is part of a multi-agent robotic system, the use of an additional light source on one agent can lead to troublesome lighting conditions (ex. blinding) for the other agents. In \cite{rgb1}, the combination of LiDAR and RGB cameras has been explored, focusing on Simultaneous Localization And Mapping (SLAM), as well as in object detection for perceptually-degraded SubT environments. Similarly, in \cite{rgb3}, the combination has been used for autonomous cooperative aerial robotics in search and rescue operations.

The use of more specialized and expensive sensors like thermal cameras can help to mitigate the negative effects of lighting in subterranean environments. In \cite{thermal3}, images from thermal cameras are used to enhance object detection and in \cite{thermal1}, and \cite{thermal5} to improve SLAM. However, solutions using thermal cameras cannot perceive features in thermal equilibrium with the SubT environment, like the reflective markers and signs, which are part of SubT environments as a standard, to aid the humans and robots to localize themselves in the absence of GPS.

In the recent years, the use of event cameras, which are equipped with a new and unconventional sensing technology based on perceiving asynchronous events as against the synchronous recording of frames, is becoming popular. The disruptive advances in event-based vision have revealed a wide range of opportunities for tackling problems that are proven to be challenging for RGB-cameras and the scope of event cameras has become broader with the increasing adoption of this technology (\cite{Event_Vision_Survey}). Recently, the fusion of event cameras and LiDARs has been pursued in \cite{Depth_Event}, \cite{Event_GUided_Depth_Sensing}, and \cite{L2E}. Nevertheless, the potential of this combination of sensors in SubT environments is unexplored and this is precisely the subject of this article.
\subsection{Contributions}
    \begin{figure}[t]
        \centering
        \includegraphics[width=\columnwidth]{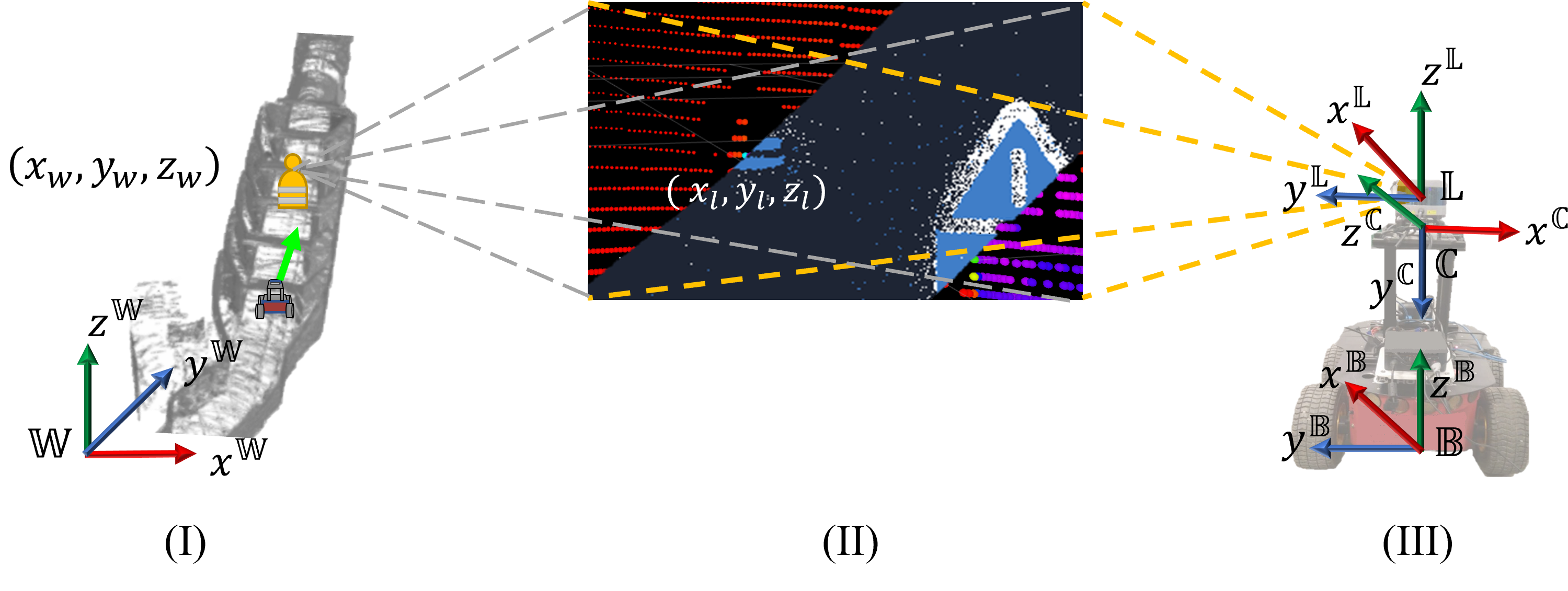}
        \caption{The proposed fusion modality is used for human and object detection in SubT environments (I). Event camera and LiDAR streams are filtered, clustered and paired for relative localization $(x_l, y_l, z_l)$ in the LiDAR's frame of reference (II). A robotic platform (Pioneer 3AT) transforms $(x_l, y_l, z_l)$ into the world frame $(x_w, y_w, z_w)$ and uses NMPC to track a human or robot in a SubT environment (III).}
        \label{fig:diag}
    \end{figure}
We propose a novel LiDAR and event camera fusion modality for fast and precise object and human detection in a wide variety of adverse lighting conditions found in SubT environments. Fig.~\ref{fig:diag} depicts a concept visualization of human detection and tracking in a SubT environment. The proposed fusion robustly extracts the locations of infrastructure features (reflective markers) commonly present in SubT environments, by exploiting a less well known capability of the event camera to see the LiDAR infrared light beams reflected from the environment, especially the high-intensity reflections from reflective markers.
%
The locations-of-interest extracted by the fusion modality are transformed into the world frame and passed as references to a Nonlinear Model Predictive Controller (NMPC) for reactive tracking of humans in SubT environments.
%
Finally, we present an experimental validation of the proposed detection and tracking scheme in a real SubT environment using a Pioneer 3AT mobile robot. The experimental results demonstrate the capability of human detection and tracking in a variety of adverse lighting conditions, such as low or no light conditions, high-contrast zones and in the presence of blinding light sources. 
%
%
%
\section{Event cameras for subterranean environments} \label{sec:SubT}
\subsection{Adverse lighting conditions in SubT environments}
Some of the most commonly present adverse lighting conditions in SubT environments are described below.
\begin{enumerate}
\item \textbf{Low light}. In most subterranean environments, the presence of natural light is low and in some cases nonexistent. 
Algorithms based on vision sensors find it hard to extract useful information from dark zones. 
%
\item \textbf{High contrast}. In a mine, it is common to find sources of lighting installed to aid human workers in a mine. However, ensuring uniform lighting across the mine is expensive. Consequently, it is common to find well-illuminated areas interspersed by non-illuminated ones. While the human eye is capable of adapting to such circumstances, most RGB-camera based solutions struggle and lose details in the darker or brighter zones.
\item \textbf{Blinding lights}. Human operators working in non-illuminated zones carry light sources, like the ones installed on mining helmets. This leads to situations where human operators and cameras are blinded to a great degree by lights from other individuals, heavy vehicles and machinery operating in mines. This reveals the need for robotics systems involved in SubT environments to be able to handle the presence of blinding lights.
\end{enumerate}

Figures \ref{fig:ex1}, \ref{fig:ex2}, and \ref{fig:ex3} present one case each for the three adverse lighting scenarios mentioned so far. The pictures were taken in a mine of the EPIROC company on the outskirts of \"{O}rebro, in southern Sweden. The equipment and reflective vests provided by EPIROC and seen in the figures are an accurate representation of the protocols and equipment normally used by personnel in mines. 
%
\subsection{Event cameras in LiDAR-illuminated environments}
Event cameras have proven to be substantially faster in detecting change in a scene in comparison to traditional RGB-cameras. Because of the recording of events corresponding to changes in the scene, event cameras can handle high dynamic range in the scene being perceived. In this article, we exploit a less well known capability of the event camera to see the infrared light beams from a LiDAR that are reflected from the environment. LiDARs illuminate the environment using an active infra-red light source and use the reflections from the environment to sense the surroundings by reconstructing a point cloud. The event camera also picks up the reflections from the LiDAR-illuminated scene, especially the reflections with high intensities returning from highly reflective markers present on vests and sign boards. 

\begin{figure}[!b]
        \centering
        \includegraphics[width=\columnwidth]{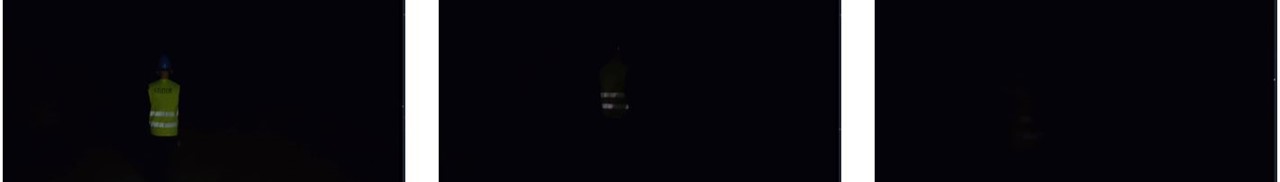}
        \caption{Low light. Example of a person walking into a dark tunnel until a point where he is no longer visible.}
        \label{fig:ex1}
    \end{figure}

\begin{figure}[!b]
        \centering
        \includegraphics[width=\columnwidth]{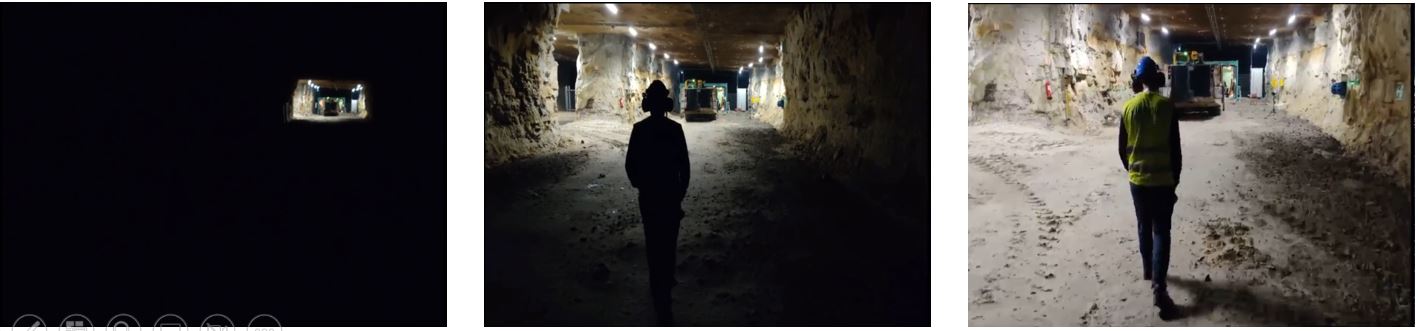}
        \caption{High contrast. Example of a person walking from a dark zone into an illuminated zone.}
        \label{fig:ex2}
    \end{figure}

\begin{figure}[!b]
        \centering
        \includegraphics[width=0.5\columnwidth]{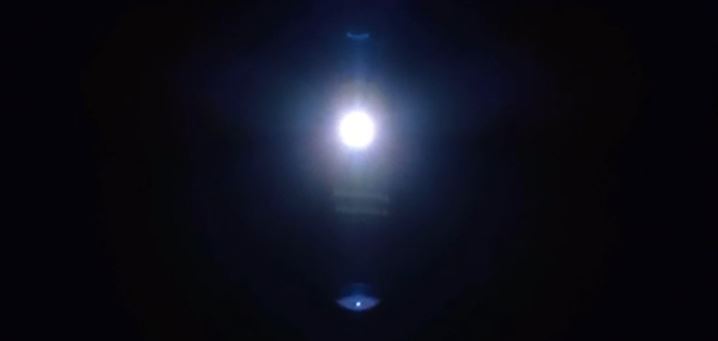}
        \caption{Blinding lights. Example of a person in a dark tunnel, flashing his helmet light at the camera.}
        \label{fig:ex3}
    \end{figure}

\begin{figure}[!b]
        \centering
        \includegraphics[width=\columnwidth]{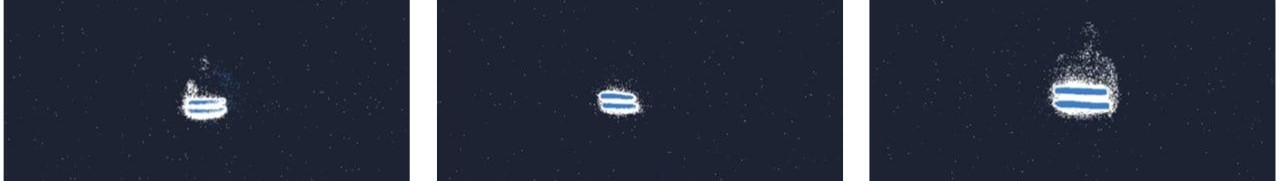}
        \caption{Low light. Event camera images in contrast to the RGB-camera images show in Fig.~\ref{fig:ex1}}
        \label{fig:ex1e}
    \end{figure}

\begin{figure}[!b]
        \centering
        \includegraphics[width=\columnwidth]{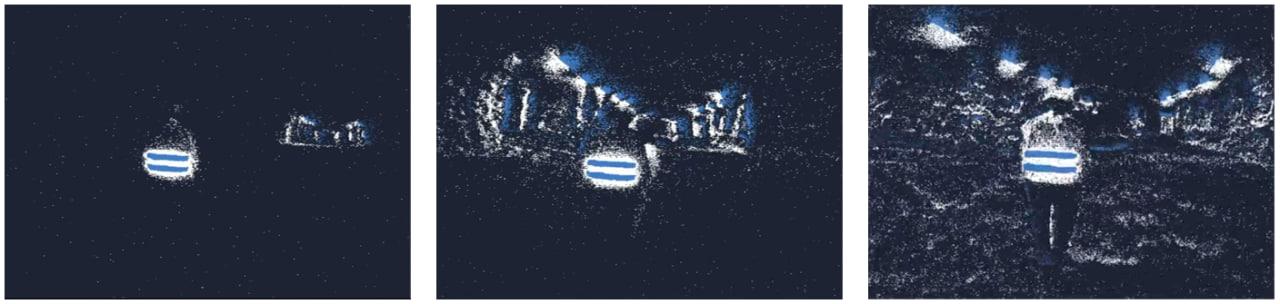}
        \caption{High contrast. Event camera images in contrast to the RGB-camera images show in Fig.~\ref{fig:ex2}}
        \label{fig:ex2e}
    \end{figure}

\begin{figure}[!b]
        \centering
        \includegraphics[width=0.5\columnwidth]{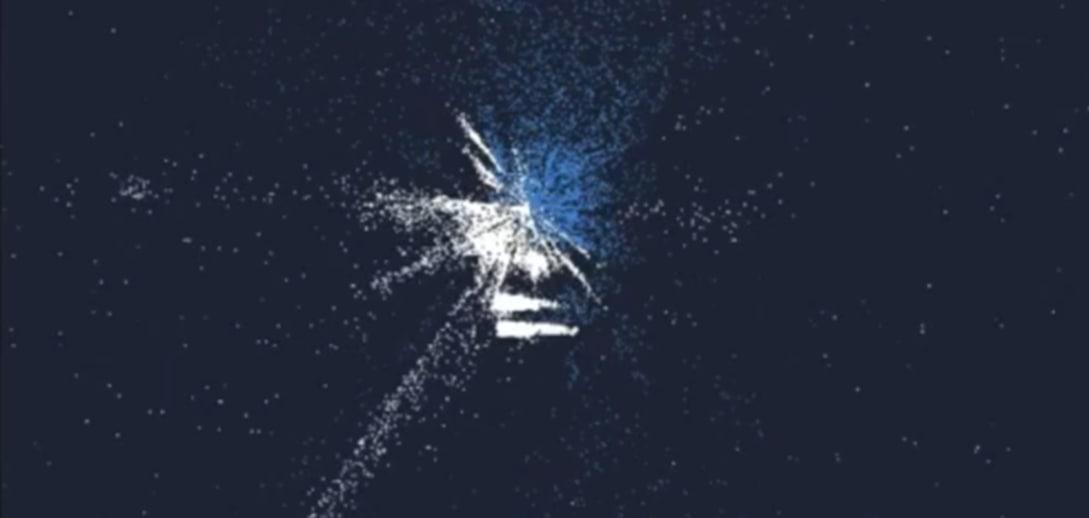}
        \caption{Blinding lights. Event camera image in contrast to the RGB-camera image show in Fig.~\ref{fig:ex3}.}
        \label{fig:ex3e}
    \end{figure}

The degree and level of detail of the sensed scene depends greatly on the relative placement of event camera with respect to a LiDAR and on the calibration of the camera. When the event camera is placed directly below the LiDAR, the event camera picks the LiDAR reflections from the environment. We use this relative configuration in the fusion proposed in this article for human detection (through the reflective markers on the vests) in the adverse scenarios presented in Figures \ref{fig:ex1}, \ref{fig:ex2}, and \ref{fig:ex3}. The view as seen by an event camera are presented in Figures \ref{fig:ex1e}, \ref{fig:ex2e}, and \ref{fig:ex3e}. 
%

The event camera takes advantage of the "blinking" nature of a LiDAR (every blink recorded as an event), thus allowing it to detect even objects in the environment that are completely still. This is a novel and certainly an unconventional use of the event camera, as the classical event camera based solutions usually seeks to distil "moving" objects from a static background. This property provides a fundamental advantage to the fusion of event cameras and LiDARs, as in the LIDAR illuminated scene, the event camera senses both static and moving objects.
\subsection{Event camera and LiDAR}
%
%

\begin{figure}[!t]
        \centering
        \begin{subfigure}[b]{0.49\columnwidth}
            \centering    \includegraphics[width=\textwidth]{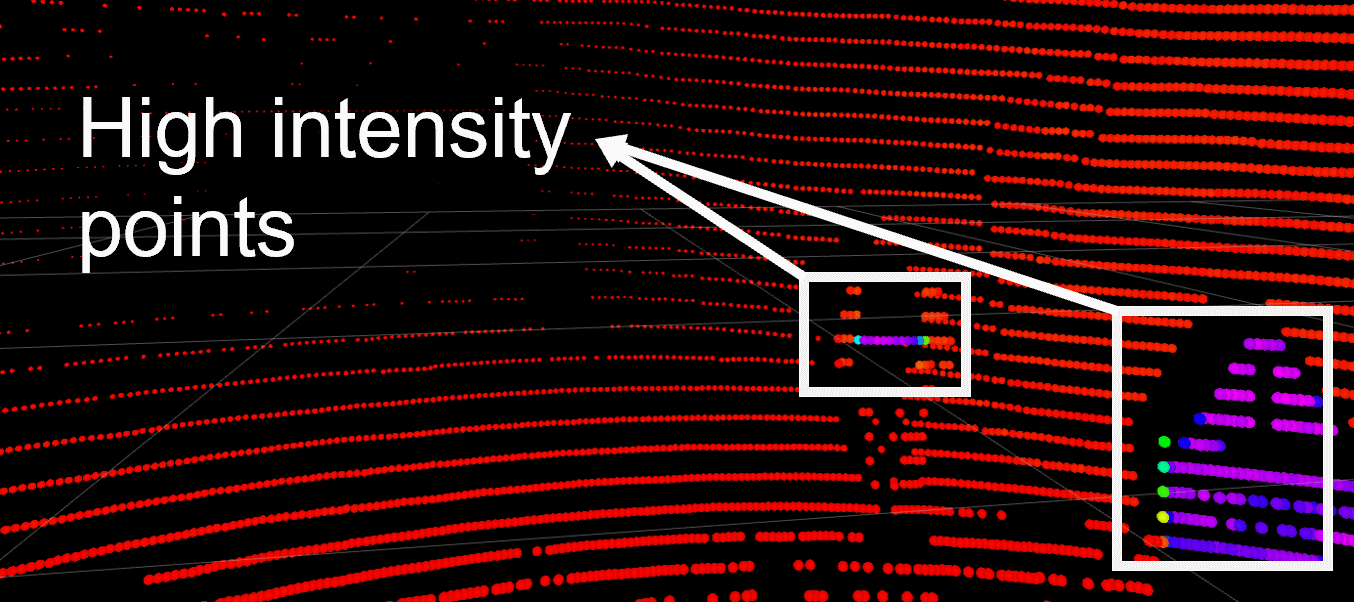}
            \caption{ }
            \label{lidar_intensity}
        \end{subfigure}
        \hfill
        \begin{subfigure}[b]{0.49\columnwidth}
            \centering \includegraphics[width=\textwidth]{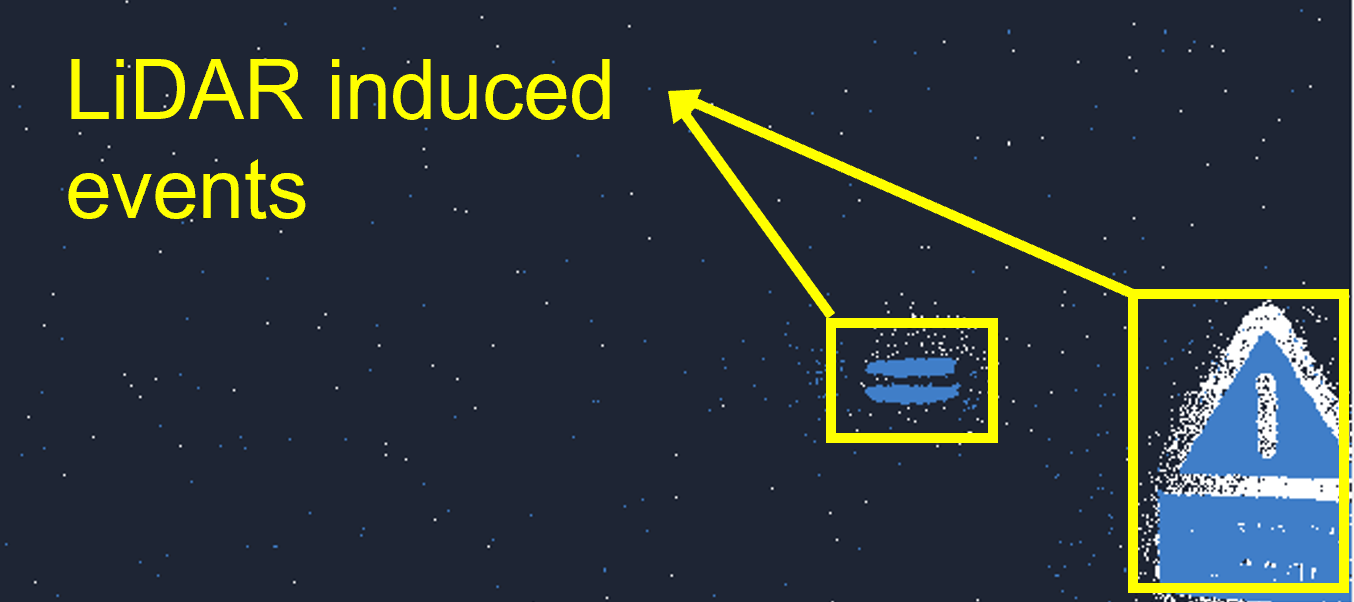}
            \caption{ }
            \label{event_intensity}
        \end{subfigure}
        \caption{A scene with a human operator and a sign board from a dark tunnel, as seen in (a) A LiDAR point cloud and (b) The event camera image.}
        \label{fig:lid2}
    \end{figure}

In this section, we elaborate on some important aspects of the LiDAR and event camera fusion. The use of event cameras alongside LiDARs offers an alternative to the lone LiDAR paradigm and allows for faster and more robust human detection. This is due to the fact that event cameras are able to detect 2D objects, allowing them to classify objects based on the shape of the detected patterns. In the scenario presented in Fig. \ref{fig:lid2}, it is easier to distinguish an arrow from the two horizontal lines of a person in an event camera feed than in a LiDAR point cloud. 
Furthermore, the point cloud from the LiDAR becomes sparse with distance and the reflections from the reflective stripes on a huilluminate the environment using an active infra-red lightman standing farther away will not return to the LiDAR. Such reflections are still visible on the event camera, and the proposed fusion modality would still keep track of objects farther away.
%
%
%

Another advantage of the LiDAR and event camera fusion is that precise and robust calibration between the sensors is not necessary. This is due to the fact that both point cloud and events can be filtered in order to leave only the points or events belonging to the reflective markers. This is precisely the strategy used in the algorithm presented in the next section. 
%
Finally, for the proposed fusion framework, an event camera does not require any other light source besides the LiDAR, nor does it introduce disturbances to the operation of the LiDAR. Furthermore, in multi-agent scenarios with more systems using a similar fusion framework, it is just a matter of using LiDARs with different operating frequencies. In this setup, frequency filtering on the feed of the event camera could be used to distinguish reflections from different LiDARs.
%
%
%
%
%
%
%
%
%
%
%
\section{Human Detection in Subterranean Environments}\label{sec:human_detection}
\begin{figure}[t]
        \centering
        \includegraphics[width=\columnwidth]{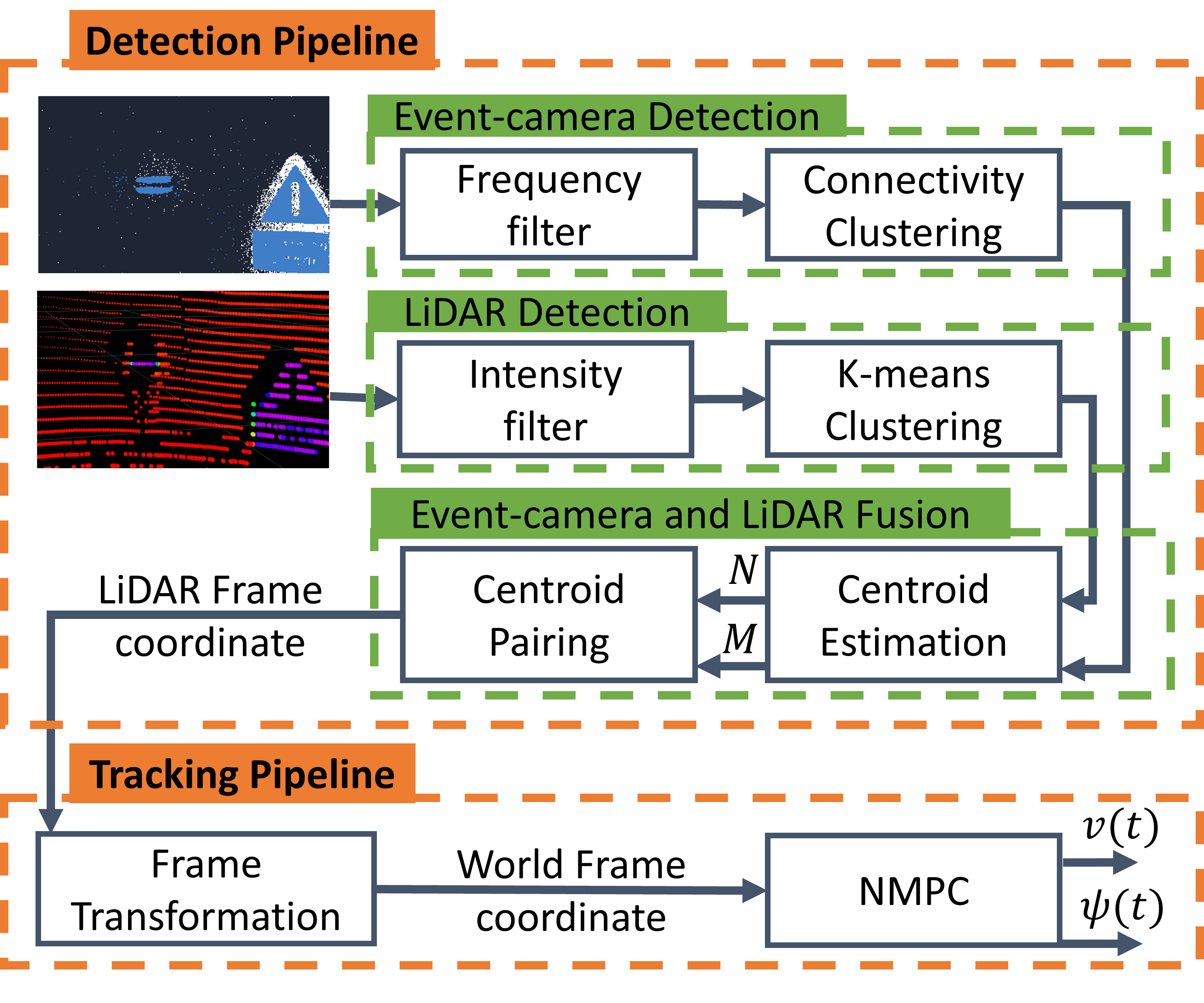}
        \caption{A block diagram representation of the LiDAR-Event Camera fusion framework proposed for  human detection and tracking.}
        \label{fig:diag2}
    \end{figure}

In this section, we present a novel human detection and tracking method for subterranean environments. The functional block diagram in Fig. \ref{fig:diag2} presents an overview of the proposed method which is divided into two main pipelines: i) Detection, presented in this section and ii) Tracking, which is detailed in section \ref{sec:human_tracking}. In event camera detection, frequency filtering followed by connectivity clustering is used. In LiDAR detection, intensity filtering followed by K-means clustering is employed. The unique mapping between the centroids of the filtered clusters from the two sensors is found, thus locating the relative positions of all humans and objects with reflective markers. A centroid of interest is then used by the tracking pipeline to set a reference for the nonlinear MPC-based human tracking algorithm.
The rest of the section provides detailed information of all the sub-components of the detection pipeline. 
\subsection{Object detection using frequency filtering on the event stream}\label{subsec:detection}
%
%
%
%
Next, we elaborate on the frequency filtering technique that is used to detect pixel locations where events occur at a specified frequency. An event from an event camera is a small information package that contains the position in image coordinates ($x_p$ and $y_p$) where the event (change in intensity) occurred, a timestamp ($t$) indicating the time of occurrence of the event and the polarity ($p$), denoting the direction of the change in intensity. In order to determine the frequency of events at a pixel, we can simply compute the difference between the current and previous timestamps. 
%
%
Pixel locations where the frequency of events falls in a specified range can be filtered, following which we cluster them together based on their connectivity in the following manner: A pixel $p \in \mathbb{R}^{2}$ with frequency $h$ belongs to a cluster $Q$, if $p$ is an 8-neighbor (either sharing an edge or a vertex) of any given pixel $q \in {Q}$ with frequency $g$ and $h \simeq g$. Then we estimate the centroid of each cluster as $C_{Qx} = \sum_{i=1}^{j} Q_{xi}$ and $C_{Qy} =  \sum_{i=1}^{j} Q_{yi}$, where $C_{Qx}$ and $C_{Qy}$ are the coordinates of the centroid of the cluster in image coordinates and $j$ is the number of pixels in the cluster. We repeat the process with each cluster to obtain $N = [(C_{Q1x},C_{Q1y}), \dots, (C_{Qnx},C_{Qny})]$, where $N$ is a list of all the cluster centroids from clusters $Q_1$ to $Q_n$. The set $N$ contains the image coordinates of different object detections from a scene, an example for which is shown in Fig.~\ref{fig:fil1}. 
%
%
\subsection{Object detection using intensity filtering on the LiDAR point cloud}
\begin{figure}[!htbp]
         \centering
         \begin{subfigure}[b]{0.49\columnwidth}
             \centering
             \includegraphics[width=\textwidth]{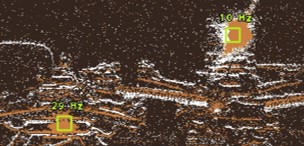}
             \caption{Event camera detection.}
         \label{fig:fil1}
         \end{subfigure}
         \hfill
         \begin{subfigure}[b]{0.49\columnwidth}
             \centering
             \includegraphics[width=\textwidth]{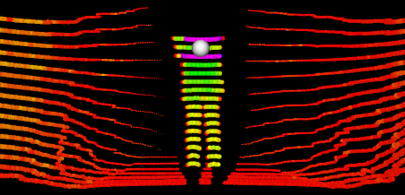}
             \caption{LiDAR detection.}
         \label{fig:fil2}
         \end{subfigure}
         \caption{ (a) The estimated centroids of the event clusters in the 
         detection of the positions of two infra-red light sources. (b) The estimated centroid of a cluster 
         in a LiDAR point cloud is shown as a white sphere.}
         \label{fig:fil}
     \end{figure}

We filter the point cloud based on the intensity of the points. As mentioned before, points belonging to the reflective markers have substantially higher intensities than the rest of the points. Once filtered, we cluster the high-intensity points using K-means clustering, with cluster size $m$. Here we select $m$ to be equal to $n+1$. This is done to ensure that when an object of interest comes out of the event camera field of view, the algorithm can still locate the corresponding cluster in the point cloud data. 
Once the different clusters in the LiDAR point cloud have been found, we proceed to estimate the centroids of the detected clusters, in order to obtain a precise 3D location of the object in the LiDAR frame as shown in Fig.~\ref{fig:fil2}. Finally, we collect all the estimated centroids of the clusters $O_1$ to $O_m$ into the set $M$, where $M =  [(C_{O1x},C_{O1y}), \dots, (C_{Omx},C_{Omy})]$.
%
%
%
\subsection{Event camera and LiDAR fusion}
\begin{figure}[!htbp]
         \centering
         \begin{subfigure}[b]{0.32\columnwidth}
             \centering
             \includegraphics[width=\textwidth]{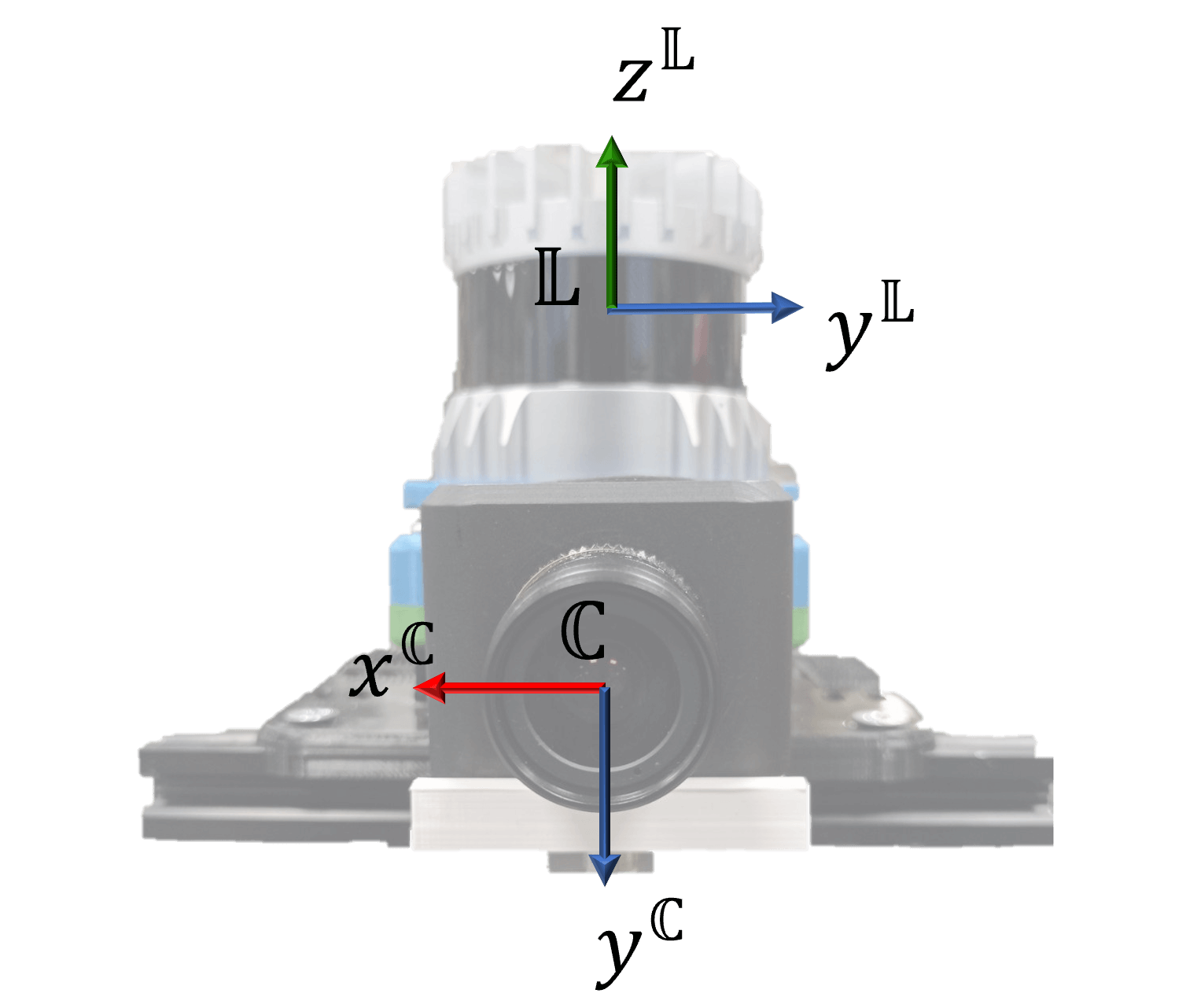}
             \caption{Front view }
         \end{subfigure}
         \hfill
         \begin{subfigure}[b]{0.32\columnwidth}
             \centering
             \includegraphics[width=\textwidth]{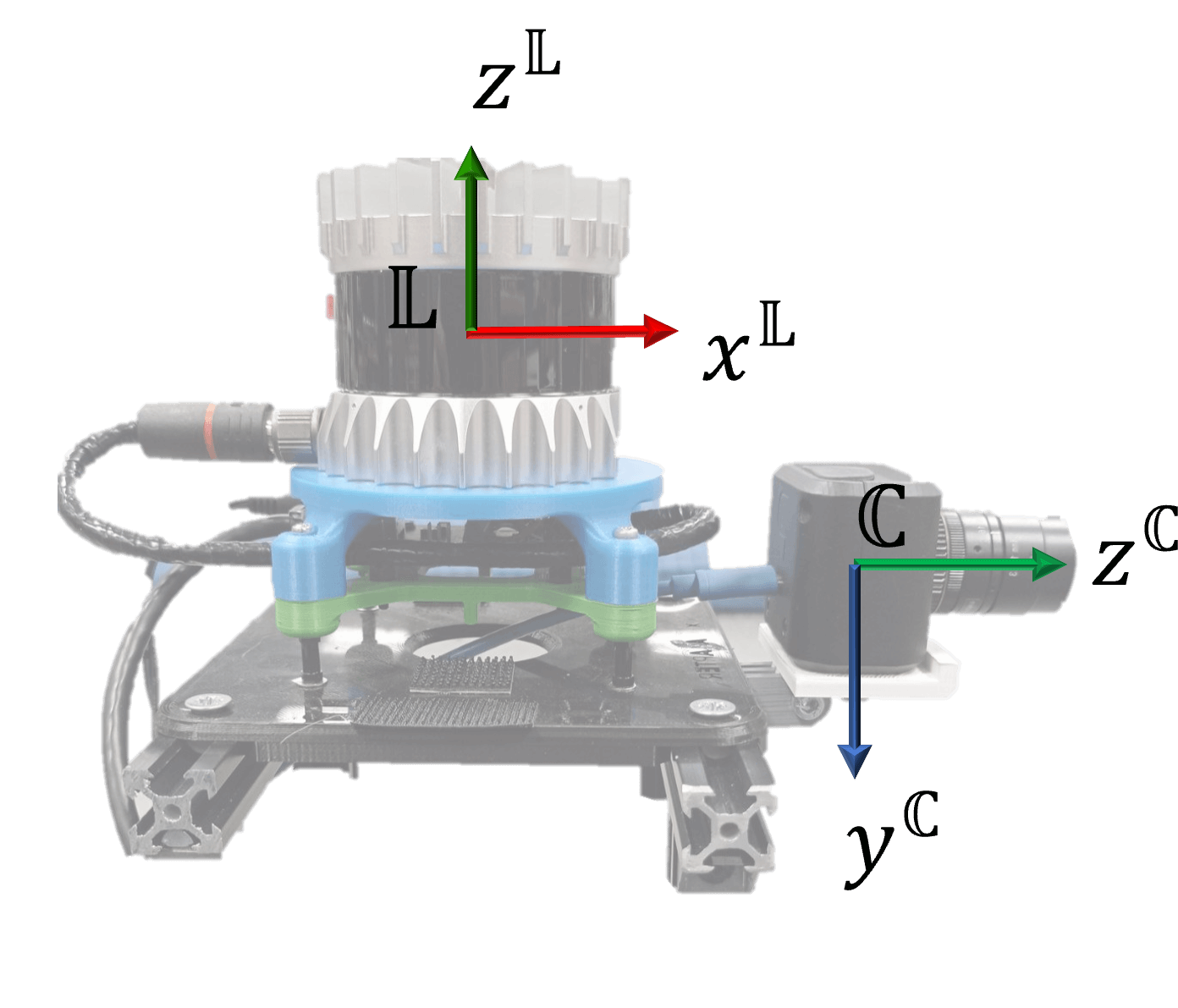}
             \caption{Side view }
         \end{subfigure}
         \hfill
         \begin{subfigure}[b]{0.32\columnwidth}
             \centering \includegraphics[width=\textwidth]{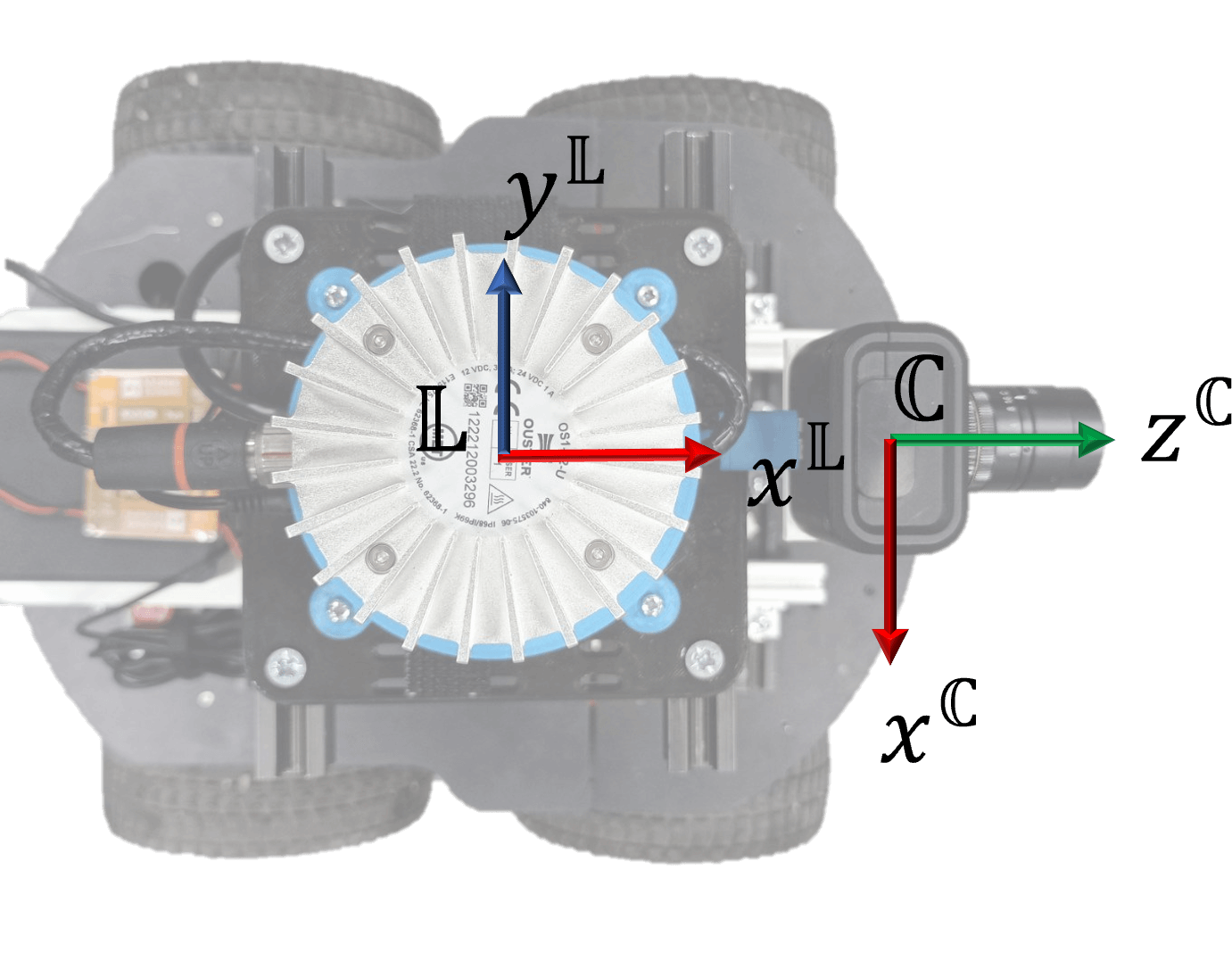}
             \caption{Top view }
         \end{subfigure}
         \caption{Relative positioning of the event camera and LiDAR with the corresponding frames-of-reference.}
         \label{fig:fra}
     \end{figure}

Having found the set of centroids $N$ and $M$, we now proceed to pair each element of the set $N$ with a unique element in the set $M$. To do so, first we estimate the angle of the centroids in $N$ and $M$, 
with respect to the front of the robotic platform, which is the same as the front of the event camera image and the LiDAR point cloud. Let $\theta_{Nu}$ and $\theta_{Mu}$ be the estimated angles of the pair of centroids $N_u$ and $M_u$ with respect to the $z^\mathbb{C}$ axis of the image frame and the $x^\mathbb{L}$ axis of the LiDAR frame respectively. A depiction of the event camera and LiDAR frames can be found in Fig.~\ref{fig:fra}.
We aim to find a pairing that minimizes the sum of the differences in angle between the two clusters by minimizing the following objective function: $\argmin \sum_{u=1}^{n} f(\theta_{Nu},\theta_{Mu})$, where $f$ is the difference between the pair of angles $\theta_{Nu}$ and $\theta_{Mu}$, i.e $f(\theta_{Nu},\theta_{Mu}) = \mid \theta_{Nu} - \theta_{Mu} \mid$. Since the centroids belonging to $N$ do not have a $z$-coordinate, we estimate their angle using the point $(N_{ux},\, N_{uy},\, 1)$ as an approximation. In the event of single object tracking, as is the case in this paper, we will end with a single centroid pair, with coordinates $(x_{l},\;y_{l},\;z_{l}) \in \mathbb{R}^{3}$ in the LiDAR frame.
\section{NMPC-based human tracking}
\label{sec:human_tracking}
In this section, we elaborate on the Nonlinear Model Predictive Control (NMPC) based tracking pipeline shown in Fig. \ref{fig:diag2}. NMPC is an optimal control technique
capable of handling the state and input constraints of the system. The NMPC controller used in this work for human tracking utilizes a differential drive mobile robot model and accepts pose references in inertial frame, in order to optimally actuate the robot such that smooth human following is achieved. 
%
%

\textbf{Lidar frame to inertial frame transformation:} In the previous section, the detection pipeline localized an object to $(x_{l},\;y_{l},\;z_{l})$ in LiDAR frame $\mathbb{L}$. For setting a reference for the NMPC based tracking algorithm, we perform the transformation from the LiDAR frame $\mathbb{L}$ to inertial frame $\mathbb{W}$. 
%
%
%
%
 Since the robotic platform is a wheeled 2D mobile robot with the LiDAR aligned along the $z$ axis of the robot, transformation of the $z$-coordinate is not necessary.
%
%
%
%

\textbf{Position tracking control:} In this work, we utilize Direct Lidar Odometry (\cite{chen2022direct}) along with IMU information to get precise position of the robot in inertial frame. This is necessary to accurately follow the references $(x_w, y_w)$ derived in the previous subsection. As presented in Fig. \ref{fig:diag2}, the proposed framework detects objects of interest in the LiDAR fixed coordinate frame and the same is transformed into the inertial coordinate frame in order to set a reference for a classical NMPC controller. The NMPC controller uses the nonlinear model of a differential-drive mobile robot for model prediction and incorporates 1) upper and lower bound constraints on the control inputs: the linear velocity  ($\nu(t)$) and  the angular velocity ($\psi(t)$) of the mobile robot and 2) a safety constraint of maintaining a distance of $1.5$m from the human being tracked. We do not elaborate on the formulation of the NMPC controller here, as it is being used in its standard form. We refer the reader to \cite{NMPC} for a detailed description.  

When executing human tracking, the sensor fusion framework encounters the following four scenarios: 
(1) Both the event camera and LiDAR find a cluster. The corresponding reference is sent to the NMPC. (2) Only the event camera finds a cluster. The angle reference is sent, with coordinates in front of the robot, until the robot gets close enough to the human and the LiDAR begins to detect a cluster. (3) Only the LiDAR finds a cluster. The corresponding angle reference is send to the NMPC so that the robot aligns with the human and the event camera begins to pick the cluster. (4) Neither the event camera nor the LiDAR find a cluster. A reference to hold current position is given to the NMPC. In all relevant cases, the tracking references are set such that a security margin of $1.5$m is maintained between the robot and the human target.
%

\section{Experimental Setup}
\label{sec:hardware}
We evaluated the performance of the proposed fusion framework for human detection and tracking in a real SubT environment using a modified version of the Pioneer 3AT mobile robot.
The robot uses the PX4 Cube low-level controller
for actuation and it is equipped with a forward-looking Prophesee event camera.
 In addition, the robot is equipped with a Ouster LiDAR OS0.
The onboard processing unit is an Intel NUC 11 Pro Mini PC
with  Intel Core i7-1185G7 Processor and 16GB DDR4 of RAM. The robot is commanded with  linear velocity ($\nu(t)$) and angular velocity ${\psi}(t)$, which are the control inputs. The experimental trials for human detection and tracking were performed in the Lule\r{a} Mjolkuddsberget mine (\cite{mjolkudsberget}) in northern Sweden. The mine includes tunnels with sections containing illumination and no illumination.
The experiment's objective was for the robot to follow a person, maintaining a distance greater than or equal to $1.5$m, in all adverse lighting conditions present in the mine. For this, the robot was made to follow a person across a series of tunnels, some of which had no illumination. The NMPC based tracking module ensured that human following was achieved while maintaining a safe distance from the human. 
%
%
%
\section{Experimental Results}
\label{sec:experiments}
  \begin{figure}[b]
        \centering
        \includegraphics[width=\columnwidth]{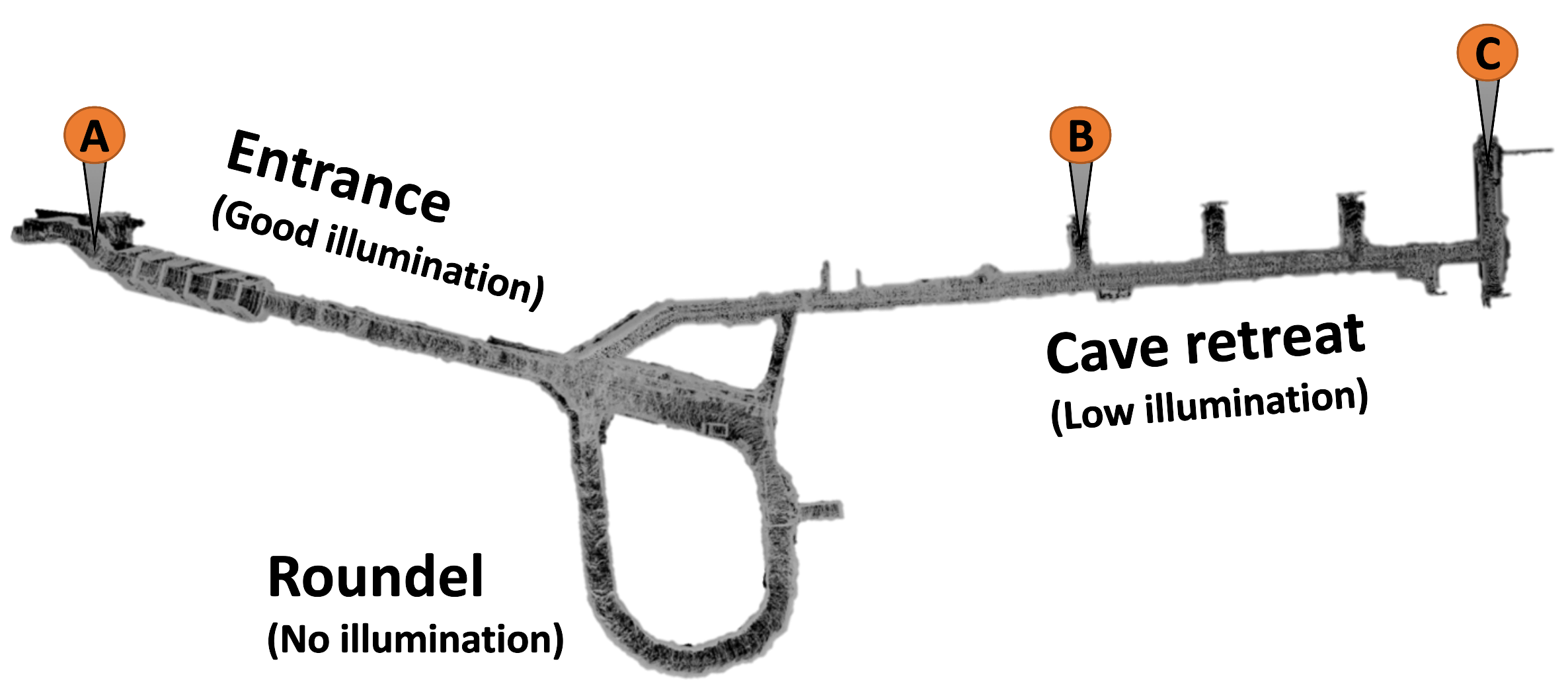}
        \caption{A map of the test site indicating the start and end points, and different sections of the mine along with the illumination conditions.}
        \label{fig:res0}
    \end{figure}

     \begin{figure}[t]
         \centering
         \begin{subfigure}[b]{0.49\columnwidth}
             \centering
             \includegraphics[width=\textwidth]{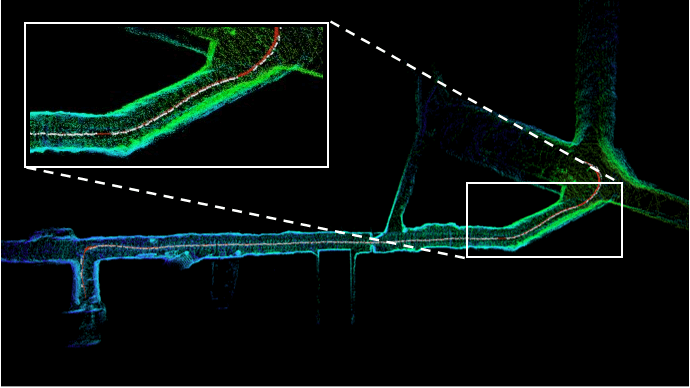}
             \caption{ Cave retreat}
         \end{subfigure}
         \hfill
         \begin{subfigure}[b]{0.49
         \columnwidth}
             \centering
             \includegraphics[width=\textwidth]{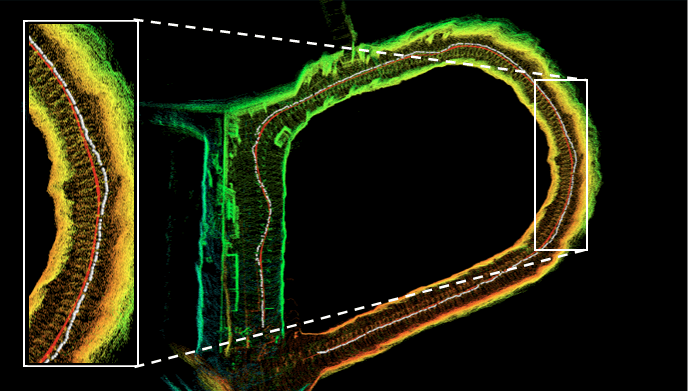}
             \caption{ Roundel}
         \end{subfigure}
         \hfill
         \begin{subfigure}[b]{\columnwidth}
             \centering \includegraphics[width=\textwidth]{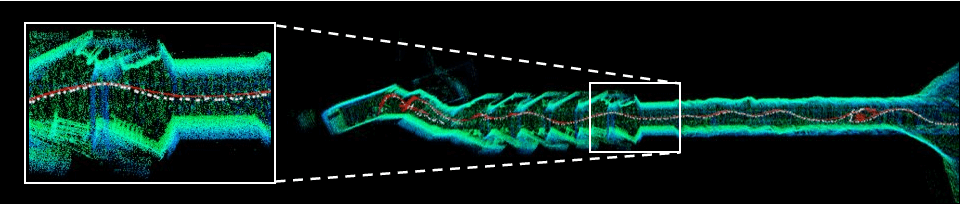}
             \caption{Entrance }
         \end{subfigure}
         \caption{The tracking performance of the robotic platform in different sections of the test environment. The positions of the human target, as estimated by the fusion framework, are indicated as white spheres. The trajectory of the robot is indicated using the red line.}
         \label{fig:res1}
     \end{figure}
%
%
%
%
%
The experiment was performed a total of three times, with different start and end points as shown in Fig.~\ref{fig:res0}. The test environment consisted of 1) \textit{Entrance}, an open and rectangular tunnel with good illumination, 2) \textit{Roundel}, circular tunnel without illumination, and 3) \textit{Cave retreat}, a long and mostly rectangular tunnel with low illumination. In the first experiment, the route from B to A (286m in 392s) was chosen. For the second experiment, the route from A to C (352m in 440s) was picked, and finally the last experiment was from C to A and then back to B (638m in 836s). In all the three experiments, the demonstration went through the roundel section, and in the last experiment, it was made to go through the roundel both the times. 
Fig.~\ref{fig:res1} presents the tracking performance of the robot in different sections of the test route. The estimated position of the human target is represented with white spheres and the trajectory followed by the robotic platform is represented with the red line. On all the runs, the robot successfully followed the human target throughout the routes, while maintaining a safety distance even in the most adverse lighting condition in the test environment. A video of the experiment can be found in the following link: \url{https://youtu.be/kIJ61VyIVTM}
%
    
%
    
    
%
\section{Conclusions}\label{sec:conclusions}
%
%
This article presented an event camera and LiDAR fusion framework for human detection and tracking in subterranean environments. The proposed framework locates humans and objects-of-interest by detecting the reflective markers commonly used in SubT environments for safe human operation. The fusion framework employed filtering and clustering on both the event camera and LiDAR streams and paired the clusters to robustly localize humans in the robot's local frame-of-reference. The detected locations are then transformed into the inertial frame to set references to a Nonlinear Model Predictive Controller for reactive human tracking.
The human detection and tracking system was deployed and experimentally validated in a real life SubT environment in Lule\r{a} Mjolkuddsberget mine. The proposed algorithm successfully handled non-illuminated and high contrast zones along the test routes, thus demonstrating fully autonomous reactive human tracking in real SubT environments. 
%
%
%
\bibliography{ifacconf} 
\end{document}